# Opinion Mining Based Entity Ranking using Fuzzy Logic Algorithmic Approach.

Pratik N. Kalamkar [1,*], A.G. Phakatkar [2]

[1]*Department of Computer Engineering, Pune Institute of Computer Technology, Pune, India.*
[2]*Department of Computer Engineering, Pune Institute of Computer Technology, Pune, India.*

**Abstract.** Opinions are central to almost all human activities and are key influencers of our behaviors. In current times due to growth of social networking website and increase in number of e-commerce site huge amount of opinions are now available on web. Given a set of evaluative statements that contain opinions (or sentiments) about an Entity, opinion mining aims to extract attributes and components of the object that have been commented on in each statement and to determine whether the comments are positive, negative or neutral. While lot of research recently has been done in field of opinion mining and some of it dealing with ranking of entities based on review or opinion set, classifying opinions into finer granularity level and then ranking entities has never been done before. In this paper method for opinion mining from statements at a deeper level of granularity is proposed. This is done by using fuzzy logic reasoning, after which entities are ranked as per this information.

Keywords: Fuzzy reasoning; Information Search and Retrieval; Query processing; Text analysis; Text mining.

## 1. Introduction

Opinion mining, also called Sentiment analysis, is the field of study that analyses people's opinions, sentiments, evaluations, appraisals, attitudes, and emotions towards entities such as products, services, organizations, individuals, issues, events, topics, and their attributes [1]. The rapid growth of the social media and internet has made available lot of opinions regarding certain services, product or any other entity freely available on internet. Opinions are central to almost all human activities and are key influencers of our behaviors. People's decisions are motivated by what other think about certain service, product. Current search engines or ranking methods work basically on information retrieval. This is very active research area. There are several reasons for this. First, it has a wide arrange of applications, almost in every domain. Second, it offers many challenging research problems, which had never been studied before.

---

\* Corresponding author
.

While lot of research recently has been done in field of opinion mining and some of it dealing with ranking of entities based on review or opinion set, classifying opinions into finer granularity level and then ranking entities has never been done before. In this paper method for opinion mining from statements at a finer level of granularity using fuzzy logic approach and then rank entities as per this information is proposed. Starting with, consider some background of some previous research in field of entity ranking and opinion mining. Then study of our proposed system in detail is done that tries to make entity ranking a better experience for user by overcoming shortcomings of earlier methods and using some already done good quality research in opinions mining.

## 2. Literature Survey : Survey of similar systems along with Pros and cons

Use of opinions in order to rank entities based on analysis of all review of the entity and user's query at a more granular level is itself challenging [10]. This is so because; Opinion mining faces challenges like irony and sarcasm. Therefore sentiments analysis is great challenge while analyzing opinions expressed. In such scenario mining opinion at a more detail level of granularity will be challenging for entity ranking.

Samaneh Nadali in his paper "Sentiment Classification of Customer Reviews Based on Fuzzy logic", 2010 IEEE. Pointed out that holistic lexicon-based does not consider the strength of each opinion i.e., whether the opinion is very strongly negative (or positive), strongly negative (or positive), moderate negative (or positive), very weakly negative (or positive) and weakly negative (or positive). He gave solution to this limitation by classifying opinions into the granularity levels (i.e. very weak, weak, moderate, very strong and strong by combining opinion words (i.e. Adverb, adjective and verb) with use of fuzzy logic. He proposed four stage methods to do this that is Fuzzification of Inputs, Membership Function Design, Fuzzy Rules Design, and Defuzzification [2]( Figure 1).

This method has advantage that consideration of adverbs and fuzzy logic approach will classify opinions into more granular level. Thus while using output of this method that opinions classified into more granular level, will help us better rank entities in proposed model when use of certain ranking algorithm is made.

In another paper titled "Opinion-Based Entity Ranking" by (Kavita Ganesan & Cheng Xiang Zhai, 2010) [4] given a user's keyword query that expresses the desired features of an entity, and then ranking of all the candidate entities based on how well opinions on these entities match the user's preferences is done. The limitation of this paper being that having not considered the granularity of opinions entity ranking can hardly get more precise. So cannot get more precise ranking.

Considering the shortcomings of each of this method and space of improvement in just before discussed paper, a new system is proposed that combines the pros of Samaneh Nadali paper with Kavita Ganesan paper to give a new system. How the newly created system behaves under a standard widely used BM25 ranking algorithm will be studies. This is described in sections which follow.

## 3. Proposed System

The proposed system is explained as follows. Ranking of entities is done by first classifying their opinions into finer granular classes and then taking total summarization of its opinion set to match best with user entered query. It contains of three steps which are described as below.

Step: 1) Classification of opinions using fuzzy logic algorithmic approach.
Step: 2) Extraction of aspects using conditional random field machine learning.
Step: 3) Raking of entities so as to best match user preference.
Description of each step in detail is as follow.

*3.1) Classification of opinions using fuzzy logic algorithmic approach*

Classical logic system works well when there is clear, absolute or mathematical truth. Like is 1+1=2? Answer can take only two values viz. yes or no. On the other hand however there are some problems whose answer may depend upon user's perception or whose output may not be clear. Fuzzy logic is a form of many-valued logic; it deals with reasoning that is approximate rather than fixed and exact. Compared to traditional binary sets (where variables may take on true or false values) fuzzy logic variables may have a truth value that ranges in degree between 0 and 1. Fuzzy logic has been extended to handle the concept of partial truth, where the truth value may range between completely true and completely false [19].

Opinions classification problem is mostly same. Opinions are classified as positive, negative or as neutral. But how much positive words in opinions can classify it as positive or how much negative words in opinion can classify it as negative? Use of fuzzy logic in such cases can help us classify opinions into more granular levels of positivity and negativity. In this step take inspiration is taken from earlier mentioned technique by Samaneh Nadali, 2010 IEEE. This method as follows following steps,

*3.1.1) Finding of opinion words from sentence:* - Finding of opinions words and adverbs so as to classify opinion's strength. Opinion words are adjective and Adverbs. Use of POS (part of speech tagger) named OpenNLP is done to mark these adjective and adverbs.

*3.1.2) Fuzzy logic system:* - As discussed earlier fuzzy logic system will be implemented. Here steps will include fuzzification of input where special degree for each of this opinion words is associated viz. like: 4 love: 5, good: 3, excellent: 6,really: 5, extremely: 9, enjoy:8, very: 5. Then a triangular membership function is designed to divide into three levels low, moderate and high. A fuzzy rule designing is done to find orientation of review. Finally the fuzzy results are converted into crisp values using a defuzzification function, as below,

$$Y^* = \frac{\int y\, \mu(Y) Y\, dy}{\int y\, \mu(y)\, dy} \quad (1)$$

Where, Y* is crisp value and μ(Y) is a membership of corresponding value y is previous result.

*3.1.3) Final output:* - Final orientation along with strength is obtained finally. This same method for finding orientation and strength of query entered by user will be followed.

*3.2) Extraction of aspects using Conditional Random Field (CRF) machine learning*

After finding orientation and strength of opinions in review next step is to propose methods for aspect extraction. Various methods for aspect extraction are used viz. Extraction based on frequent nouns and noun phrases, Extraction by exploiting opinion and target relations, Extraction using supervised learning, Extraction using topic modelling. Out of this in our proposed system use the one based on supervised learning is used. And within supervised learning use of the Conditional Random Fields (CRF), hence used as short CRF at some places, a probabilistic model, to find aspects in opinions is made. These are chosen because it will help us to have a learning component, which will make improvements constantly as more data is trained. On the other hand if use of hard core methods like those based on frequent nouns, scope of improvement will be limited. Conditional Random Fields for segmenting and labelling sequence data were first proposed by John Lafferty; these models are probabilistic in nature and can be defined as follow.

Consider,
X is a random variable over data sequences to be labelled, and Y is a random variable over corresponding label sequences. All components $Y_i$ of Y are assumed to range over a finite label alphabet Y. CRF can be defined as, Let G = (V,E) be a graph such that Y = $(Y_v)$v∈V , so that Y is indexed by the vertices of G. Then (X,Y) is a conditional random field in case, when conditioned on X, the random variables $Y_v$ obey the Markov property with respect to the graph: $p(Y_v |X,Y_w,w != v) = p(Y_v |X,Y_w, w \sim v)$, where w ~ v means that w and v are neighbours in G [18].

Using above explanation of CRF extraction of the aspect from opinion is done. For doing so a training data set is provided i.e. opinions consisting of desired aspects that are to be extracted from test data and train CRF so that aspects are pointed out from review and also its syntactic dependence is resolved to know what is user's opinion regarding certain aspect. CRF also offer several advantages over HMM.

*3.3) Raking of entities so as to best match user preference*
The next and most important step of our proposed model is ranking of entities. This is so because hence before research on ranking of entities using opinions that has been done has not considered opinions strength for ranking entities. For this purpose BM25 ranking algorithm is studied. Experiment results of our previous steps i.e. opinions strength and orientation over the following algorithm will be studied. However it should be remembered that before ranking entities using the following algorithm rough group of entities based on its reviews summation will be made,

*3.3.1) Match of Aspect its Orientation and strength* - Here Entities whose review set matches in both aspect presence and its orientation and strength with user query will be ranked higher than following conditions.

*3.3.2) Match of Aspect its Orientation* - Here Entities whose review set matches in aspect and it's orientation with user query will be ranked higher than following but lower than above.

*3.3.3) Match of Aspect* - Here Entities who's review set contain only aspect that is also in user query but is not matched in its orientation and strength , will be ranked lowest.

The algorithm experimented with is BM25.Which is explained as below,
*BM25*: - BM25 is chosen because it is effective and robust for many tasks. In BM25 basically a score of an opinion set of entity and query is given by (Definition),

$$S_{bm25}(D; Q) = \sum_{t \in Q \cap D} \frac{k1 c(t,D)}{c(t,D) + k1(1 - b + b \times |D|/|\bar{D}|)} \times \log \frac{n+1}{nt} \quad (2)$$

*Opinion Mining Based Entity Ranking using Fuzzy Logic Algorithmic Approach.*

where c(t;D) and c(t;Q) are the count of term t in document D and query Q, respectively, |D| is the length of document D, |D [ | is the average document length in the collection, nt is the number of documents containing term t, and b, k1, and k3 are parameters.

Let $A_1 = \{a_{11}, a_{12}, a_{13},...a_{1n}\}$
Be the set of Aspects for Entity $e_1$
Let $E = \{e_1, e_2, e_3,....e_n\}$
Be the set of entities to be ranked (entity –> issue/product/service etc.)
For each Entity $e_1$ let $R_1$ be the set of reviews = $\{r_{11}, r_{12}, ... , r_{1n}\}$

Then the strength and orientation of all aspects $A_1$ of entity $e_1$ in $R_1$ using fuzzy logic approach as discussed earlier is found. Each Aspect will have a duplet defining it in review {orientation, strength}. User will give Query say Q, Will contain Aspects, and form for every aspect duplet {orientation, strength}. Compare Orientation and Strength of Aspect in user query Q to best match Orientation and Strength of Aspects in $A_1$ of entity $e_1$ in $R_1$. Each entity will be given score according to this match, say High score for matching both aspect orientation and strength. Moderate Score for matching aspect orientation, and low score for matching only aspect. Entities will then be ranked in descending order of score, and will be roughly grouped into these three categories. After which a BM25 ranking algorithm will be applied on them to get finally ranked list.

*Mathematical Model:-*
Mathematical model of above proposed system,
$S = \{I, F, O, S_u, F_l\}$
Where, I = Input set = $\{E, R_i, Q\}$
$E = \{e_1, e_2, e_3,....e_n\}$ be the set of entities to be ranked
$R_i = \{r_{i1}, r_{i2},....r_{in}\}$ be the set of reviews for entity $e_i$
Q = user query
$F = \{f_{ext}, f_{os}, f_{sum}, f_{rank}\}$
*Let*, $A_i = \{a_{i1}, a_{i2}, a_{i3}, ....,a_{in}\}$ be the set of Aspects for Entity $e_i$
$f_{ext}$ = Function to extract aspects Ai from Review set Ri, and user query Q
$f_{ext}(Ri) = A_{ext}$
$A_{ext} = \{a_{i1}, a_{i2}, a_{i3}, ....a_{in}\}$ be the set of extracted entities.
$f_{os}$ = function to find orientation and strength of Ri for aspect $A_i$ and from user query
$f_{os}(Ai) = \{orientation, strength\}$
$f_{sum}$ = function to find overall effective orientation and strength of Ri for aspect Ai
$f_{sum} = \sum f_{os}(Ai)$
$f_{rank}$ = function to order entity set E
$O = \{e_1, e_2, e_3,...,e_n\}$ i.e. output set = Ranked set of Entity
Su = $\{e1 > e2 > e3 >...> en\}$ i.e. entities ranked as per user query preference.
Fl = Entities not properly ranked as per query preference

1450

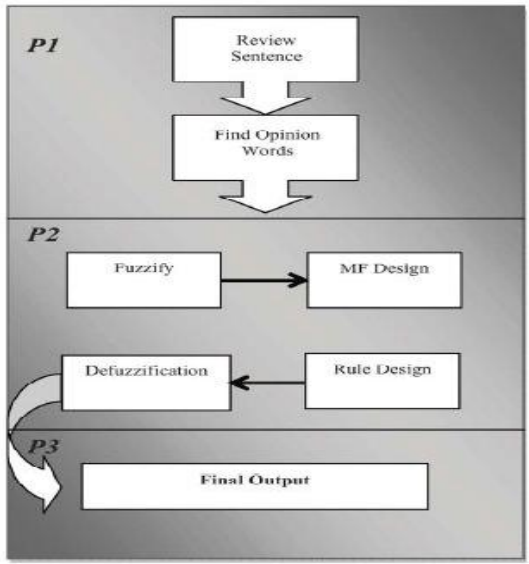
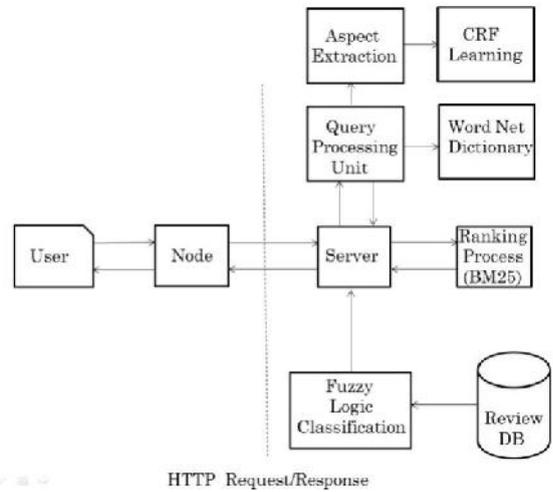

Fig. 1. System architecture Diagram   Fig.2. System by Samaneh Nadali, 2010 IEEE [2]

*Identification of Morphism (function mapping):-*

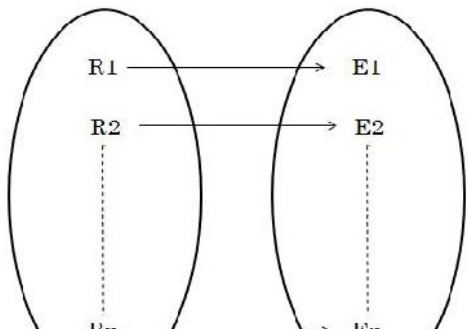
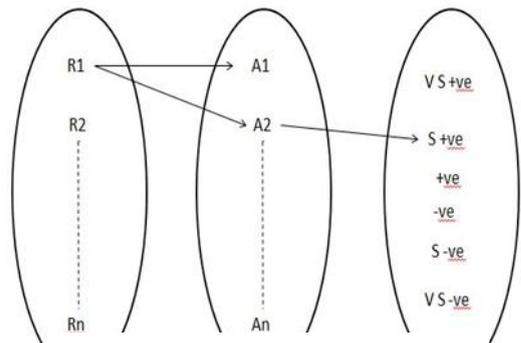

Fig.4. Set of Reviews and Entities.    Fig.3. Reviews, Aspect and Sentiments.

The overall mentioned above method can be explained with help of example below.

*Opinion Mining Based Entity Ranking using Fuzzy Logic Algorithmic Approach.*

Suppose user wants to find a Laptop that has good battery life and clear display. For this user this two aspects viz. battery and display are more important than other aspect of a laptop. In this example, laptops of various models will be our entity set. Now review set of each of this model will be analyzed using above methodology. And will be ranked so as to give user on the top of result the models which received most positive reviews about battery and display. Use of fuzzy logic to classify review and then ranking, which is used in proposed methodology, will thus help get better results than all existing systems.

## 4. Dataset and Results

The above proposed method is to be tested on a database of hotels that has over 250000 reviews of hotels from about 10 cities all over the world and on Car review set. User will enter query, and will get best matching hotel based on this query. Aspects were extracted by using CRF algorithm, for this purpose initially review dataset of 2500 reviews was taken, out of which 750 reviews were manually annotated for aspects in this semi-supervised approach. Various aspects relating to hotels such as location, TV channels, hygiene, and staff were trained and extracted. For Cars set extracted aspect were mileage, brakes, drive etc.

Comparison of ranked list result with those of normal ranked list where fuzzy logic is not being used is done. And result set is then compared.

## 5. Conclusion and Future Work

Use of these methods will greatly enhance the ranking of entity based of the review the entity gains and the user query. More precise Ranking than normal information Retrieval is obtained. Use of aspects along with its orientation and strength makes user get precise results.Use of fuzzy logic classifies opinions into more granular level.Ranking based on how well entities aspect satisfy users query.The proposed system can be extended as an add-on to for a normal search engine like Goggle, Bing etc. This will help user get more precise and crisp search results, improving usability of a search engine. Also system can be used at online shopping websites to give user better experience of ranked entities as per his entered query.

**Acknowledgment**

Authors are thankful to Prof. Potdar, HOD, Computer Department, PICT for his support and encouragement for given project. We are also thankful to Samaneh Nadadali and Kavita Ganeshan whose previous work provided foundation for our research project. At last we would like to thank PICT ME department staff for their support.